\documentclass[runningheads]{llncs}
\usepackage[T1]{fontenc}
\usepackage{graphicx}
\usepackage{amsmath}
\usepackage{subfig}
\usepackage{subfloat}
\usepackage[export]{adjustbox}
\usepackage{algorithm}
\usepackage{arevmath} 
\usepackage[noend]{algpseudocode}

\usepackage[backend=biber,style=numeric,sorting=none]{biblatex}
\begin{document}
\title{On-Device Learning with Binary Neural Networks}
%
%
\author{Lorenzo Vorabbi\inst{1,2}\orcidID{0000-0002-4634-2044} \and
Davide Maltoni\inst{2}\orcidID{0000-0002-6329-6756} \and
Stefano Santi\inst{1}}
%
\authorrunning{L. Vorabbi et al.}
%
\institute{Datalogic Labs, Bologna 40012, IT \and
University of Bologna, DISI, Cesena Campus, Cesena 47521, IT
\email{\{lorenzo.vorabbi2,davide.maltoni\}@unibo.it}\\
\email{\{lorenzo.vorabbi,stefano.santi\}@datalogic.com}}
%
\maketitle              
\begin{abstract}
Existing Continual Learning (CL) solutions only partially address the constraints on power, memory and computation of the deep learning models when deployed on low-power embedded CPUs. In this paper, we propose a CL solution that embraces the recent advancements in CL field and the efficiency of the Binary Neural Networks (BNN), that use 1-bit for weights and activations to efficiently execute deep learning models. We propose a hybrid quantization of CWR* (an effective CL approach) that considers differently forward and backward pass in order to retain more precision during gradient update step and at the same time minimizing the latency overhead. The choice of a binary network as backbone is essential to meet the constraints of low power devices and, to the best of authors' knowledge, this is the first attempt to prove on-device learning with BNN. The experimental validation carried out confirms the validity and the suitability of the proposed method.

\keywords{Binary Neural Networks  \and On-device Learning \and Continual Learning.}
\end{abstract}
\section{Introduction}

Integrating a deep learning model into an embedded system can be a challenging task for two main reasons: the model may not fit into the embedded system memory and, the time efficiency may not satisfy the application requirements. A number of light architectures have been proposed to mitigate these problems (MobileNets \cite{howard2017mobilenets}, EfficientNets \cite{tan2019efficientnet}, NASNets \cite{cai2018proxylessnas}) but they heavily rely on floating point computation which is not always available (or efficient) on tiny devices. Binary Neural Networks (BNN), where a single bit is used to encode weights and activations, emerged as an interesting approach to speed up the model inference relying on packed bitwise operations \cite{qin2020binary}. However, almost no literature work addresses the problem of training (or tuning) such models on-device, a task which is still more complex than inference because:

\begin{itemize}
\item quantization is known to affect back propagation and weights update
\item popular inference engines (e.g. Tensorflow Lite, pytorch mobile, ecc.) do not support model training
\end{itemize}

This work proposes on-device learning of BNN to enable continual learning of a pre-trained model. We start from CWR* \cite{lomonaco2020rehearsal}, a simple but effective continual learning approach that limits weight updates to the output head, and designs an ad-hoc quantization approach that preserves most of the accuracy with respect to a floating point implementation. We prove that several state of the art BNN models can be used in conjunction with our approach to achieve good performance on classical continual learning dataset/benchmarks such as CORe50 \cite{lomonaco2017core50}, CIFAR10 \cite{krizhevsky2009learning} and CIFAR100 \cite{krizhevsky2009learning}.

\section{Related Literature}

\subsection{Continual Learning}
\label{continual_learning_section}

The classical deep learning approach is to train a model on a large batch of data and then freeze it before deployment on edge devices; this does not allow adapting the model to a changing environment where new classes (NC scenario) or new items/variation of known classes (NI scenario) can appear over time. Collecting new data and periodically retraining a model from scratch is not efficient and sometime not possible because of privacy, so the CL approach is to adapt an existing model by using only new data. Unfortunately, this is prone to forgetting old knowledge, and specific techniques are necessary to balance the model stability and plasticity. For a survey of existing CL methods see \cite{de2021continual}.

In this work we focus on Single Object Recognition task addressing the two CL scenarios of NI and NC; in both cases, the learning phase of the model is usually splitted in \textit{experiences}, each one containing different training samples belonging or not to known classes (this depends on the CL scenario).
\\
CWR* mantains two sets of weights for the output classification layer: $cw$ are the consolidated weights used during inference while $tw$ are the temporary weights that are iteratively updated during back-propagation. $cw$ are initialized to $0$ before the first batch and then updated according to Algorithm \ref{algorithm_cwr} (for more details see \cite{lomonaco2020rehearsal}), while $tw$ are reset to $0$ before each training mini-batch. CWR*, for each already encountered class (of current training batch), reloads the consolidated weights $cw$ at the beginning of each training batch and, during the consolidation step, adopts a weighted sum based on the number of the training samples encountered in the past batches and those of current batch. The consolidation step has a negligible overhead and can be quantized adopting the same quantization scheme used for CWR* weights. In CWR*, during the first training experience (supposed to be executed offline) all the layers of the model are trained but from the second experience, only the weights of the output classification layer are adjusted during the back-prop stage, to simulate a real case scenario (lines ~\ref{cwr_algo:line:b1} - ~\ref{cwr_algo:line:b2} of Algorithm \ref{algorithm_cwr}).

\begin{algorithm}
\caption{CWR* pseudocode: $\overline{\Theta}$ are the class-shared parameters. Both $tw$ and $cw$ refer to the same layer index $k$ of the model.}
\label{algorithm_cwr}
\begin{algorithmic}[1]

\Procedure{CWR*}{}
\State $cw_{k}=0$ \Comment{\small$k$ is the index of the classification layer}
\State $past=0$\Comment{number of samples for each class $i$ encountered}
\State init $\overline{\Theta}$ random or from pre-trained model
\For{each training batch $B_{j}$}\Comment{$B_{j}$ is the mini-batch of index $j$}
	\State expand layer $k$ with neurons for the new classes in $B_{j}$ never seen before
	\State $tw_{k}\left [ i \right ] = \left\{\begin{matrix}
cw_{k}\left [ i \right ], & \text{if class } i \text{ in } B_{j}
\\ 
0, & \text{otherwise}
\end{matrix}\right.$
	\State train the model with SGD
	\If{$B_{j} = B_{1}$} \label{cwr_algo:line:b1}
		\State learn both $\overline{\Theta}$ and $tw_{k}$ \label{cwr_algo:line:b3}
	\Else
		\State learn $tw_{k}$ while keeping $\overline{\Theta}$ fixed \label{cwr_algo:line:b2}
	\EndIf

\For{each class $i$ in $B_{j}$}\Comment{consolidation step}
	\State $wpast_{i} = \sqrt{\frac{past_{i}}{cur_{i}}}$, where $cur_{i}$ is the number of patterns of class $i$ in $B_{j}$
	\State $cw_{k}\left [ i \right ] = \frac{cw_{k}\left [ i \right ]\cdot wpast_{i} + \left(tw_{k}\left[i \right]- avg\left(tw_{k} \right) \right)}{wpast_{i}+1}$
	\State $past_{i} = past_{i} + cur_{i}$
\EndFor
\State test the model by using $\overline{\Theta}$ and $cw_{k}$
\EndFor

\EndProcedure
\end{algorithmic}
\end{algorithm}

\subsection{Binary Neural Networks}

Quantization is a technique that yields compact models compared to their floating-point counterparts, by representing the network weights and activations with very low precision. The most extreme quantization is binarization, where data can only have two possible values, namely $-1\left(0 \right)$ or $+1\left(1 \right)$. By representing weights and activations using only 1-bit, the resulting memory footprint of the model is dramatically reduced and also the heavy matrix multiplication operations can be replaced with light-weighted bitwise XNOR operations and Bitcount operations. According to \cite{bannink2021larq}, that compared the speedups of binary layers w.r.t. the 8-bit quantized and floating point  layers, a binary implementation can achieve a lower inference time from $9$ to $12 \times$ on a low power ARM CPU. Therefore, Binary Neural Networks combine many hardware friendly properties including memory saving, power efficiency and significant acceleration; for some network topologies, BNN can be executed on device without the usage of floating-point operations \cite{vorabbi2023optimizing} simplifying the deployment on ASIC or FPGA hardware. For a survey on binary neural networks see \cite{qin2020binary}. 

\section{On-Device CWR Optimization}

\subsection{Gradients Computation}

In this section we make explicit the weights update in the classification layer; without loss of generality, a neural network $M\left ( \cdot \right )$ is composed by a sequence of $k$ layers represented as:

\begin{equation}
M\left ( \cdot \right ) = f_{w_{k}}\left ( f_{w_{k-1}}\left ( \cdots f_{w_{2}}\left( f_{w_{1}}\left( \cdot \right ) \right ) \right ) \right )
\end{equation}

where $w_{i}$ represents the weights of the $i^{th}$ layer. In CWR* the temporary weights $tw_{k}$ (lines \ref{cwr_algo:line:b3} and \ref{cwr_algo:line:b2} of Alg. \ref{algorithm_cwr}) are updated according to Equations \ref{weight_update_formula} and \ref{weight_update_formula_2}, whose quantization is discussed in the next section.
 Denoting with $a_{i}$ and $a_{i+1}$ \footnote{Note that the output $a_{i+1}$ of level $i$ corresponds to the input of level $i+1$} the input and output activations of the $i^{th}$ layer respectively, with $\mathcal{L}$ the loss function, the backpropagation process consists in the computation of two different sets of gradients: $\frac{\partial \mathcal{L}}{\partial a_{i}}$ and $\frac{\partial \mathcal{L}}{\partial w_{i}}$.

In CWR* the on-device backpropagation algorithm is limited to the last layer which can be considered a linear layer (with a non-linear activation function) with the following forward formula: 
\begin{equation}
\label{linear_layer_formula}
a_{k+1} = f_{k} \left( o_{k+1} \right), \;\; o_{k+1} = a_{k}W_{k} + b_{k}
\end{equation}

where $a_{k+1}$ represents the output of the neural network.

Considering a classification task (with \textit{M} classes) with an unitary batch size, the \textit{Cross-Entropy} loss function is formulated as:

\begin{equation}
\label{cross_entropy}
\mathcal{H}\left(y,\; a_{k+1} \right) = -\sum_{i=0}^{M-1} y^{i}log\left(a^{i}_{k+1}\right)
\end{equation}

where $y^{i}$ represents the element of an one-hot encoded vector of ground truth and $a^{i}_{k+1}$ is the $i^{th}$ output activation sample. Using the softmax as activation for the last layer, reported below:

\begin{equation}
\label{softmax_formula}
a_{k+1}\left(o_{k+1}^{t} \right) = \frac{e^{o_{k+1}^{t}}}{\sum_{j=1}^{M} e^{o_{k+1}^{j}} }
\end{equation}

, the gradient formulas for the last classification layer can be expressed using the chain rule:

\begin{align}
\label{gradient_formula_chain_rule}
\frac{\partial \mathcal{H}}{\partial W_{k}} &= \frac{\partial \mathcal{H}}{\partial a_{k+1}} \frac{\partial a_{k+1}}{\partial o_{k+1}}\frac{\partial o_{k+1}}{\partial W_{k}} \\
\frac{\partial \mathcal{H}}{\partial b_{k}} &= \frac{\partial \mathcal{H}}{\partial a_{k+1}} \frac{\partial a_{k+1}}{\partial o_{k+1}}\frac{\partial o_{k+1}}{\partial b_{k}}
\end{align}

The final expression for Eq. \ref{gradient_formula_chain_rule} using the Eq. \ref{cross_entropy} as loss function and \ref{softmax_formula} as non-linear $f_{k}\left( \cdot \right)$ is a well-known result, that can be easily derived:

\begin{align}
\label{weight_gradient_formula}
\frac{\partial \mathcal{H}}{\partial W_{k}} &= \left( a_{k+1} - y \right) a_{k} \\
\label{weight_gradient_formula_2}
\frac{\partial \mathcal{H}}{\partial b_{k}} &= \left( a_{k+1} - y \right)
\end{align}

Using a stochastic gradient descent optimizer with learning rate $\eta$, the weights update equation is:

\begin{align}
\label{weight_update_formula}
W_{k}^{i+1} &= W_{k}^{i} - \eta \left( a_{k+1} - y \right) a_{k} \\
\label{weight_update_formula_2}
b_{k}^{i+1} &= b_{k}^{i} - \eta \left( a_{k+1} - y \right)
\end{align}

Therefore in CWR* the temporary weights $tw_{k}$ (lines \ref{cwr_algo:line:b3} and \ref{cwr_algo:line:b2} of Alg. \ref{algorithm_cwr}) are updated according to Equations \ref{weight_update_formula} and \ref{weight_update_formula_2}, whose quantization is discussed in the next section.

\subsection{Quantization Strategy}
\label{quantization_strategy}

\begin{figure}[!h]
\centering
\includegraphics[width=0.85\linewidth]{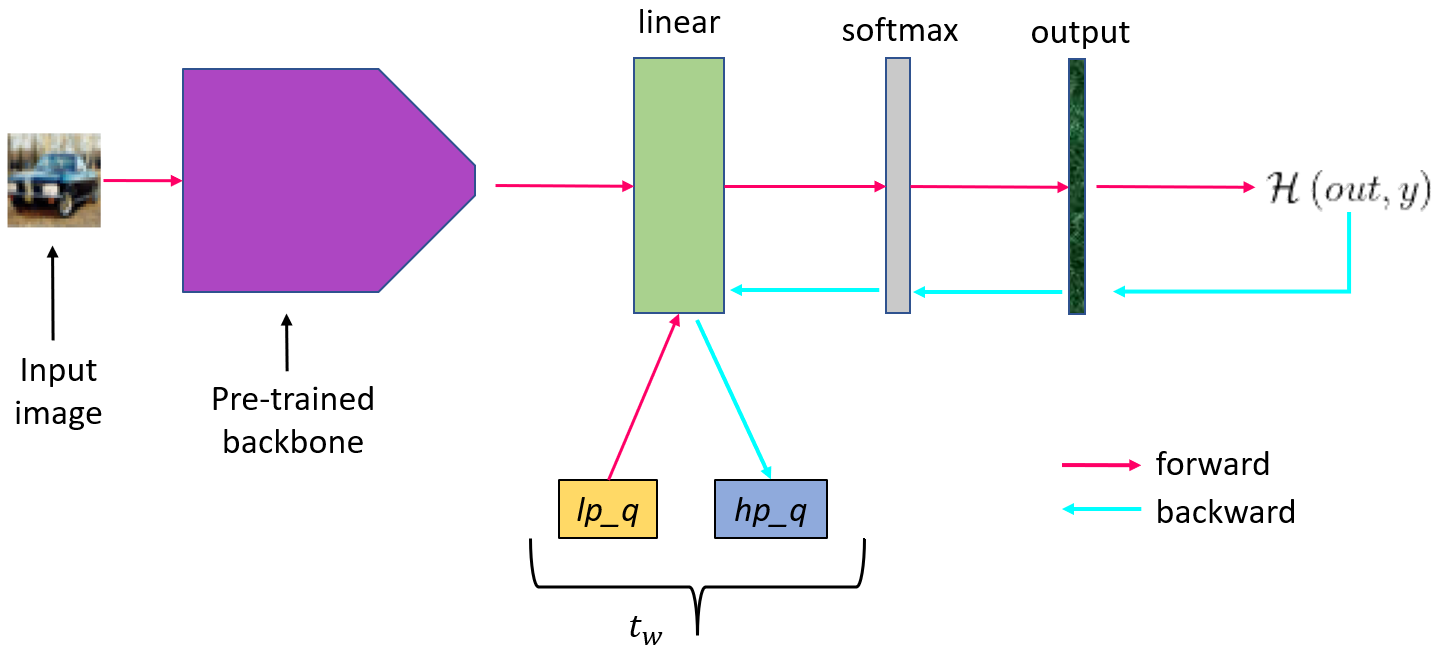}
  	\caption{Double quantization scheme that uses a different quantization level for weights/activations used in forward and backward pass.}
  	\label{fig:double_quantization_weights}
\end{figure}

Our approach considers two different quantizations: the former uses 1-bit (also called binarization) to represent weights and activations employed by the pre-trained backbone; the latter is used in the last classification layer, to quantize both forward and backward operations. This solution both reduces the latency and simplifies the adaptation of the model on new item/classes encountered.
\\ 
In particular, for the last layer quantization we followed the scheme proposed in \cite{jacob2018quantization} and implemented in GEMMLOWP library \cite{jacob2017gemmlowp}. The quantized output of a 32-bit floating point linear layer, reported in Eq. \ref{linear_layer_formula}, can be represented as:

\begin{equation}
\label{quantization_expression}
\overline{o^{int \textunderscore q}_{k+1}} = cast \textunderscore to \textunderscore int \textunderscore q \lfloor s_{k}^{int \textunderscore 32} \left( \overline{W^{int \textunderscore q}_{k}} a_{k}^{int \textunderscore q} + \overline{b_{k}^{int \textunderscore q}} \right) \rceil
\end{equation}

\begin{figure}[!t]
\centering
\includegraphics[width=0.85\linewidth]{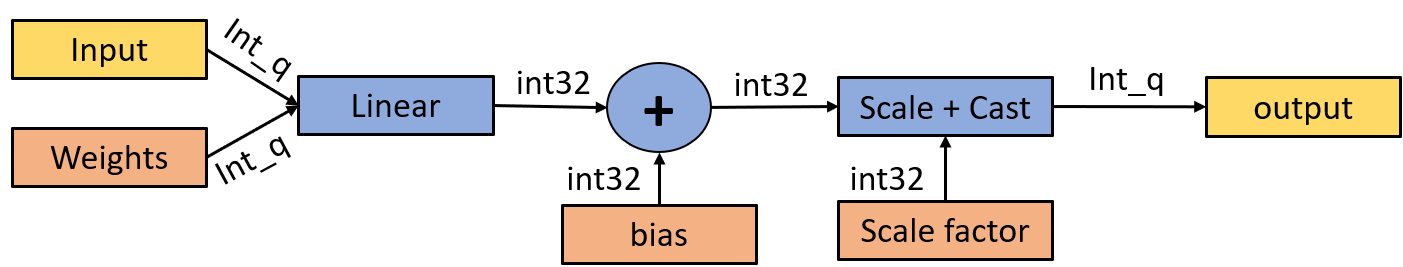}
  	\caption{Quantization scheme adopted using $q$ bits for weights and activations.}
  	\label{fig:tensor_quantization}
\end{figure}

The quantization Eq. \ref{quantization_expression} depends on the number of the quantization \textit{q} bits used $\left(8,16,32 \right)$, $\overline{\cdot}$ represents the quantized version of a tensor and $s^{int \textunderscore 32}$ is the fixed-point scaling factor having 32-bit precision, as shown in Fig. \ref{fig:tensor_quantization}. Similarly to previous works \cite{hubara2016binarized, mishra2017wrpn}, we used the straight-through estimator (STE) approach to approximate differentiation through discrete variables; STE represents a simple and hardware-friendly method to deal with the computation of the derivative of discrete variables that are zero almost everywhere.

Based on the results reported in \cite{gupta2015deep, das2018mixed, banner2018scalable}, the quantization of the gradients in Equations \ref{weight_gradient_formula} and \ref{weight_gradient_formula_2} represents the main cause of accuracy degradation during training and therefore we propose to use two separate versions of layer weights $W_{k}$, one with low-precision ($lp \textunderscore q$) and another with higher precision ($hp \textunderscore q$). As shown in Fig. \ref{fig:double_quantization_weights}, the idea is to use the $lp \textunderscore q$ version of the weights for the computations that have strict timing deadlines (forward pass), while the $hp \textunderscore q$ version is adopted during the weight update step (Equations \ref{weight_update_formula} and \ref{weight_update_formula_2}), which has typically more relaxed timing constraints (it can be executed also as a background process). Every time a new high-precision copy of weights is computed, a lower version is derived from it and stored.

\begin{figure}
	\centering
	\makebox[0pt]{%
	\subfloat[][\textit{quicknet}]
	{
	\includegraphics[width=0.65\textwidth,valign=M]{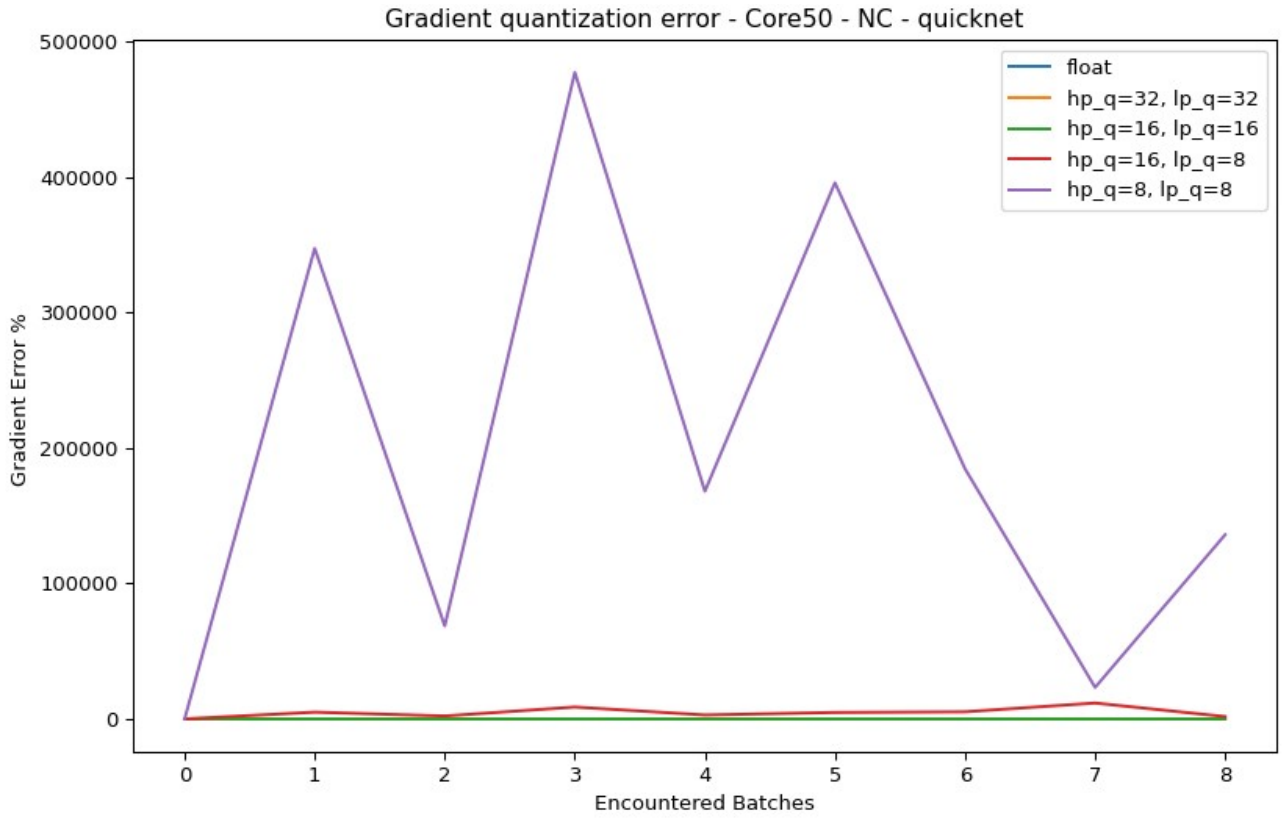}
	\label{fig:quicknet_gradient_error}
	}
	\hspace{0.5cm}
\subfloat[][\textit{realtobinary}]
	{
	\includegraphics[width=0.65\textwidth,valign=M]{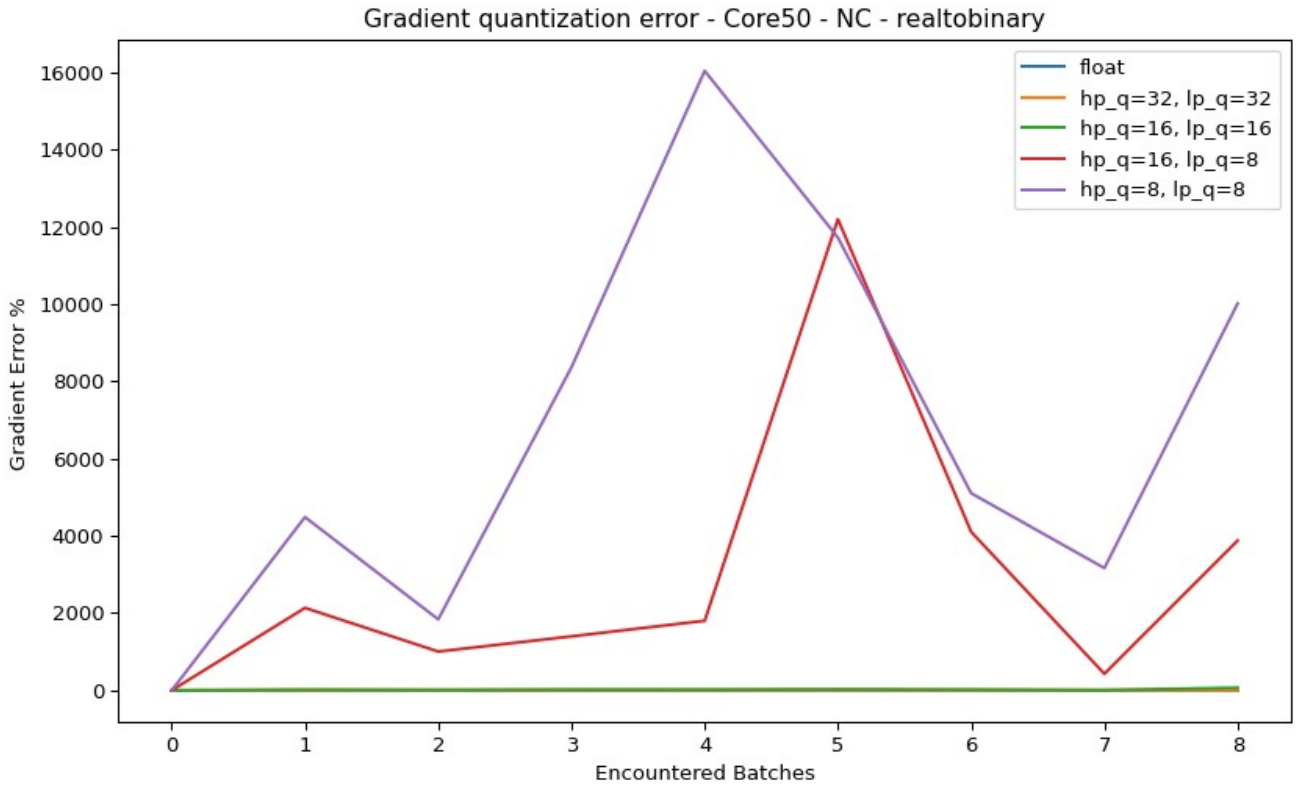}
	\label{fig:realtobinary_gradient_error}
	}}
  	\caption{Accumulation of gradient quantization errors (Mean Absolute Error in percentage) between quantized and floating-point versions for each experience. During the first experience the gradient computation is always executed in floating-point.}
  	\label{fig:quantization_error}
\end{figure}

Gradient quantization inevitably introduces an approximation error that can affect the accuracy of the model; to check the amount of approximation for different quantization levels, for each mini-batch, we compute the Mean Absolute Error (MAE, in percentage) between the floating point gradient and the quantized one for the weight tensor of the CWR* layer (for the dataset CORe50 \cite{lomonaco2017core50}). The MAE is then accumulated for all training mini-batches of each experience, as shown in Fig. \ref{fig:quantization_error}. In order to evaluate only the quantization error introduced, both floating-point and quantized gradients are computed starting from the same $W_{k}^{i}$ weights (Eq. \ref{weight_update_formula}). The plot curves of Figures \ref{fig:quicknet_gradient_error}  and \ref{fig:realtobinary_gradient_error} refer respectively to the \textit{quicknet} \cite{bannink2021larq} and \textit{realtobinary} \cite{martinez2020training} models; it is evident that the quantization error introduced using the $lp \textunderscore q$ with 8 bits is much larger compared to higher quantization schemes (16/32 bits or floating point) whose gap w.r.t. the floating point implementation is quite low, as pointed out in Section \ref{resuls_label}.

\section{Experiments}

We evaluate the proposed approach on three classification datasets: CORe50, CIFAR10 and CIFAR100 with different BNN architectures. The BNN models employed  for CORe50 have been pre-trained on ImageNet \cite{ILSVRC15} and taken from Larq repository\footnote{\url{https://docs.larq.dev/zoo/api/sota/}}; instead, the models used for CIFAR10 and CIFAR100 have been pre-trained on Tiny Imagenet\footnote{\url{http://cs231n.stanford.edu/tiny-imagenet-200.zip}}. For each dataset, we conducted several tests using a different number of quantization bits with the same training procedure. Our work is targeting a model that could continuously learn and therefore we limited the number of epochs to $10$ for the first experience and to $5$ for the remaining. The results of Eq. \ref{weight_gradient_formula} and \ref{weight_update_formula} require the adoption of the Cross Entropy as loss function and the Stochastic Gradient Descent (SGD) as optimizer; the choice of SGD is encouraged as it requires a simple computation with a limited overhead compared to the Adam \cite{kingma2014adam} optimizer. The binarization of weights and activations always happens at training time using an approximation of the gradient (STE introduced in Section \ref{quantization_strategy} or derived solutions that are model dependent) for \textit{sign} function.

Hereafter we provide some details on the datasets and related CL protocols:

\begin{description}
\item[CORe50 \cite{lomonaco2017core50}] \hspace{0.05\textwidth} It is a dataset specifically designed for Continuous Object Recognition containing a collection of $50$ domestic objects belonging to $10$ categories. The dataset has been collected in $11$ distinct sessions ($8$ indoor and $3$ outdoor) characterized by different backgrounds and lighting. For the continuous learning scenarios (NI, NC) we use the same test set composed of sessions $\#3$, $\#7$ and $\#10$. The remaining $8$ sessions are split in batches and provided sequentially during training obtaining $9$ experiences for NC scenario and $8$ for NI. No augmentation procedure has been implemented since the dataset already contains enough variability in terms of rotations, flips and brightness variation. The input RGB image is standardized and rescaled to the size of $128 \times 128 \times 3$.
\item[CIFAR10 and CIFAR100 \cite{krizhevsky2009learning}] \hspace{0.05\textwidth} Due to the lower number of classes, the NC scenario for CIFAR10 contains $5$ experiences (adding $2$ classes for each experience) while $10$ are used for CIFAR100. For both datasets the NI scenario is composed by $10$ experiences. Similar to CORe50, the test set does not change over the experiences. The RGB images are scaled to the interval $\left [-1.0\: ;+1.0  \right ]$ and the following data augmentation was used: zero padding of $4$ pixels for each size, a random $32 \times 32$ crop and a random horizontal flip. No augmentation is used at test time. 
\end{description}

On CORe50 dataset, we evaluated the three binary models reported below:

\begin{description}
\item[Realtobinary \cite{martinez2020training}] \hspace{0.05\textwidth} This network proposes a real-to-binary attention matching mechanism that aims to match spatial attention maps computed at the output of the binary and real-valued convolutions. In addition, the authors proposed to use the real-valued activations of the binary network before the binarization of the next layer to compute scaling factors, used to rescale the activations produced after the application of the binary convolution.
\item[Quicknet and QuicknetLarge\cite{bannink2021larq}] \hspace{0.05\textwidth} This network follows the previous works \cite{liu2018bi, bethge2019back, martinez2020training} proposing a sequence of blocks, each one with a different number of binary $3 \times 3$ convolutions and residual connections over each layer. Transition blocks between each residual section halve the spatial resolution and increase the filter count. QuicknetLarge employs more blocks and feature maps to increase accuracy.
\end{description}

For CIFAR10 and CIFAR100 datasets, whose input resolution is $32\times32$, we evaluated the following networks (pre-trained on Tiny Imagenet):

\begin{description}
\item[BiRealNet\cite{liu2018bi}] \hspace{0.05\textwidth} It is a modified version of classical ResNet that proposes to preserve the real activations before the sign function to increase the representational capability of the 1-bit CNN, through a simple shortcut. Bi-RealNet adopts a tight approximation to the derivative of the non-differentiable sign function with respect to activation and a magnitude-aware gradient to update weight parameters. We used the instance of the network that uses \textit{18-layers}\footnote{Refer to the following \url{https://github.com/liuzechun/Bi-Real-net} repository for all the details.}.
\item[ReactNet\cite{liu2020reactnet}] \hspace{0.05\textwidth} To further compress compact networks, this model constructs a baseline based on MobileNetV1 \cite{howard2017mobilenets} and add shortcut to bypass every 1-bit convolutional layer that has the same number of input and output channels. The $3 \times 3$ depth-wise and the $1 \times 1$ point-wise convolutional blocks of MobileNet are replaced by the $3 \times 3$ and $1 \times 1$ vanilla convolutions in parallel with shortcuts in React Net\footnote{Refer to the following \url{https://github.com/liuzechun/ReActNet} repository for all the details.}. As for Bi-Real Net, we tested the version of React Net that uses  \textit{18-layers}.
\end{description}

\begin{figure}[!t]
	\centering
	\subfloat[][Quicknet]
	{
		\includegraphics[width=0.55\textwidth,valign=M]{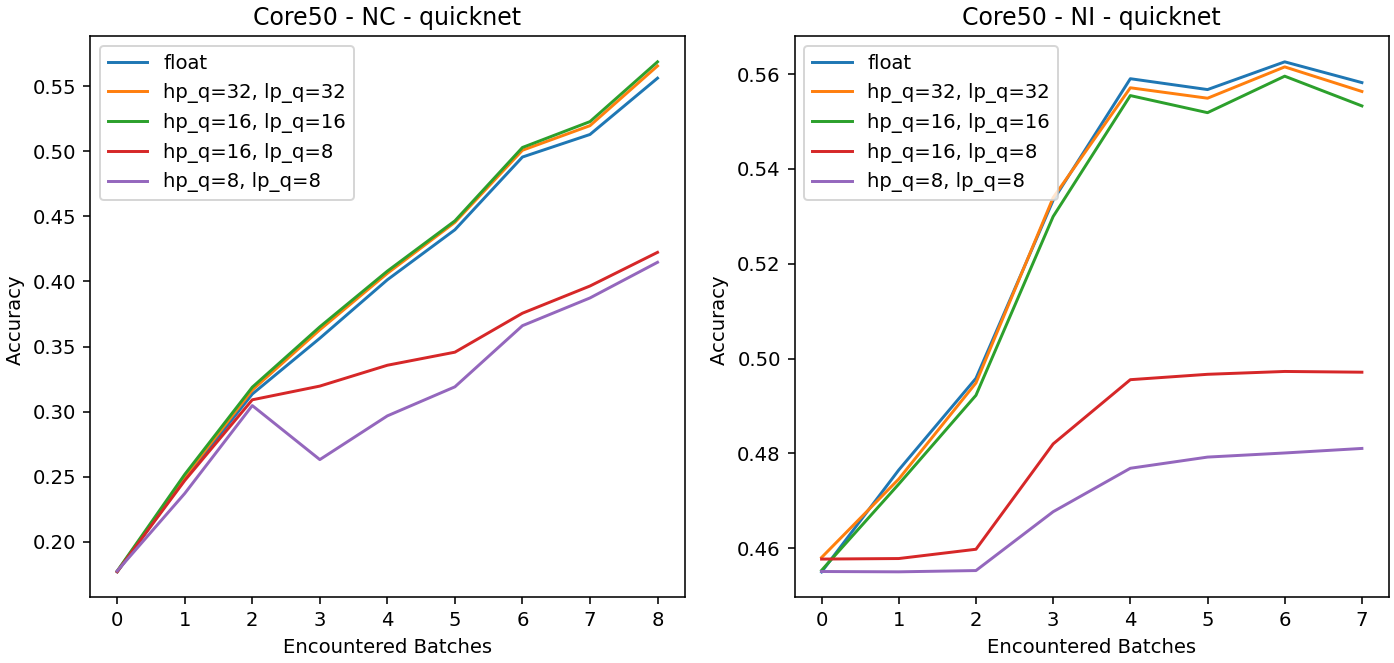}
		\label{fig:quicknet_core50}
	}
	\vfill
	\subfloat[][QuicknetLarge]
	{
		\includegraphics[width=0.55\textwidth,valign=M]{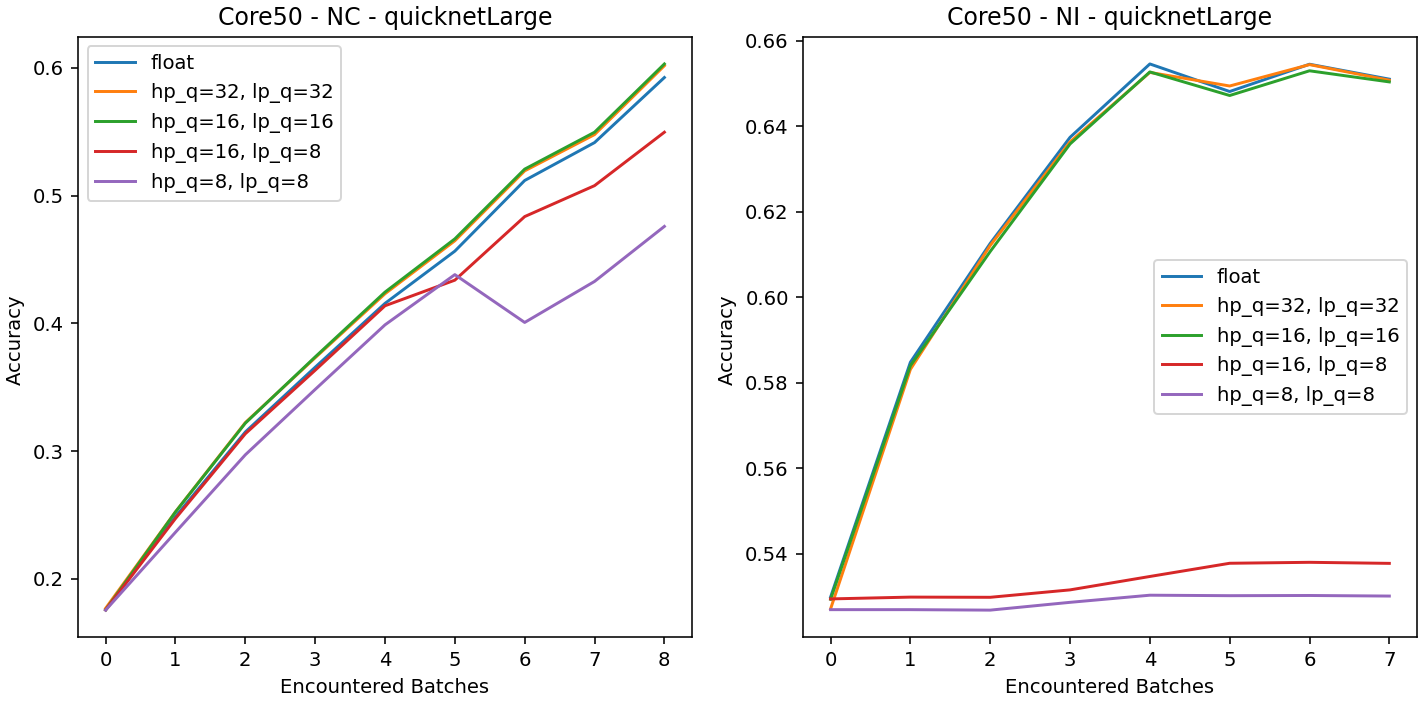}
		\label{fig:quicknetLarge_core50}
	}
	\vfill
	\subfloat[][Realtobinary]
	{
		\includegraphics[width=0.55\textwidth,valign=M]{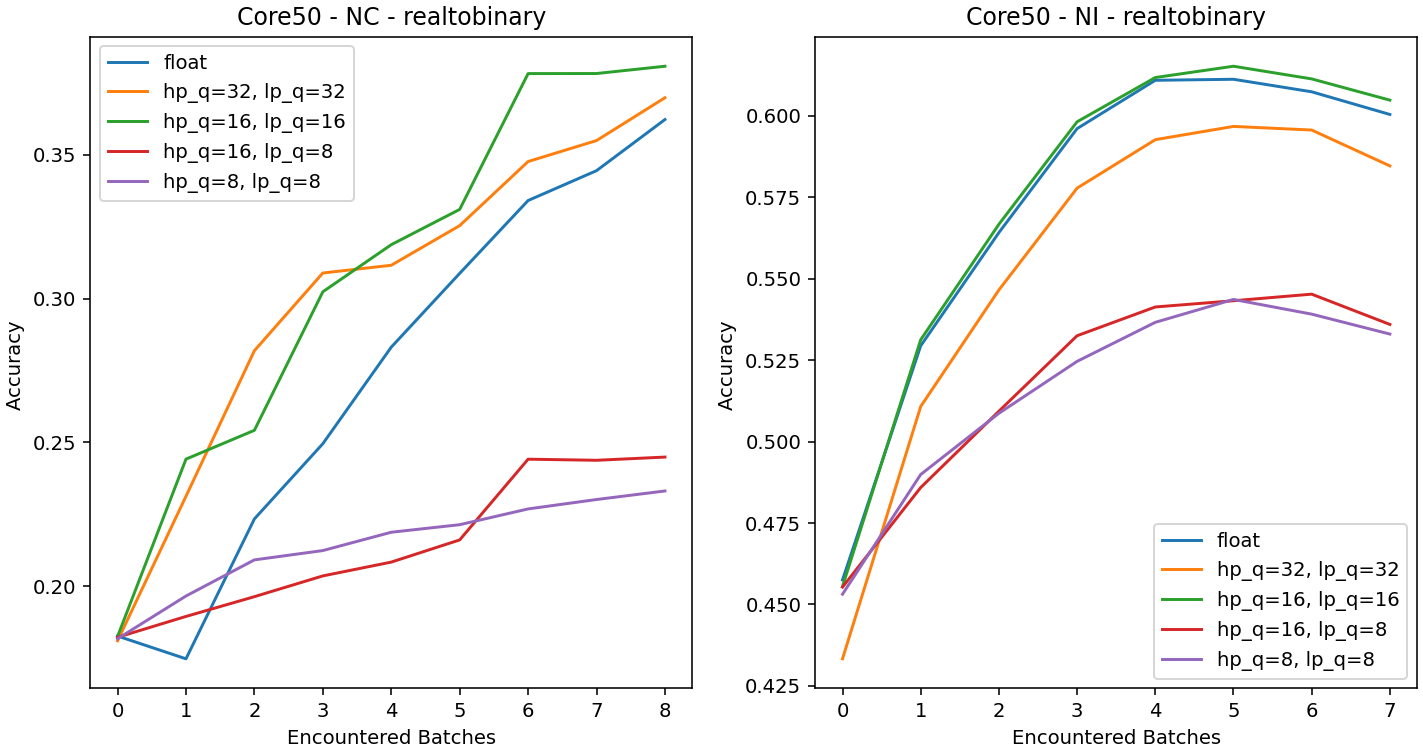}
		\label{fig:realtobinary_core50}
	}
  	\caption{CORe50 accuracy results using different quantization methods.}
    \label{Core50_accuracy}
\end{figure}

\begin{figure}
	\centering
	\subfloat[][Reactnet]
	{
		\includegraphics[width=0.55\textwidth]{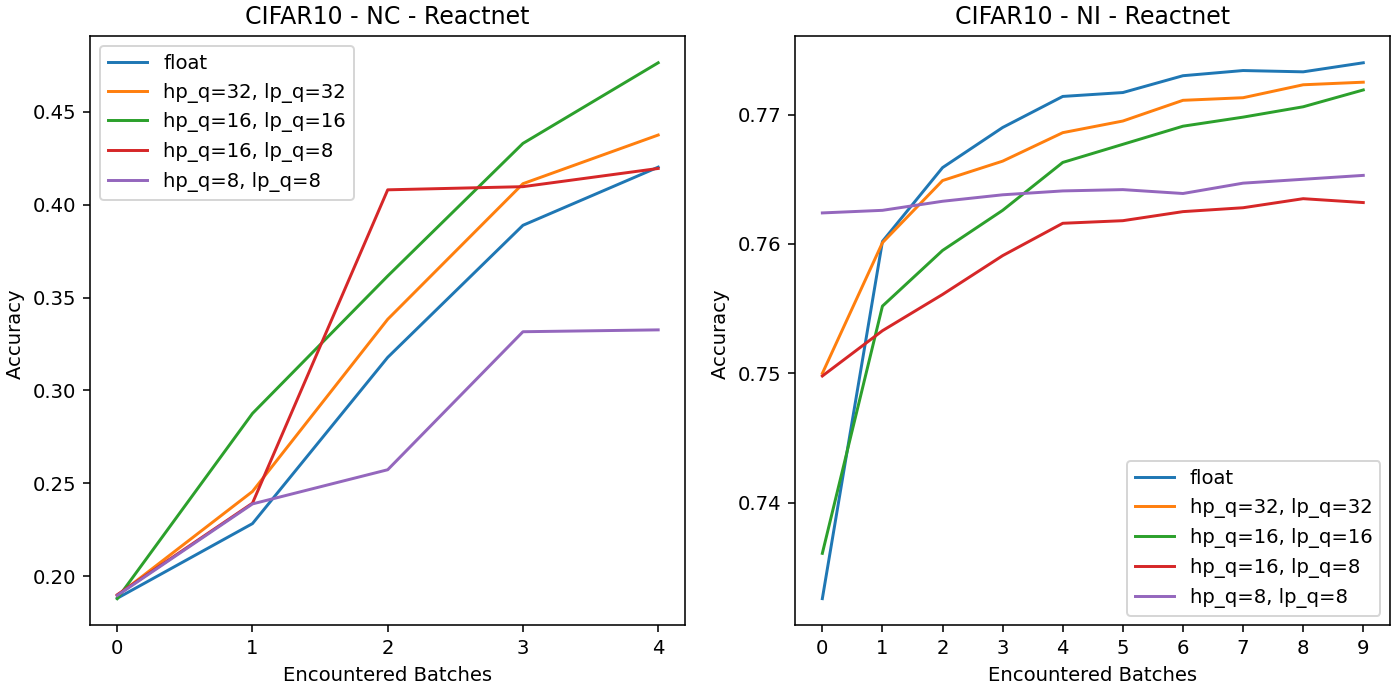}
		\label{fig:reactnet_cifar10}
	}
	\vfill
	\subfloat[][Bi-Realnet]
	{
		\includegraphics[width=0.55\textwidth]{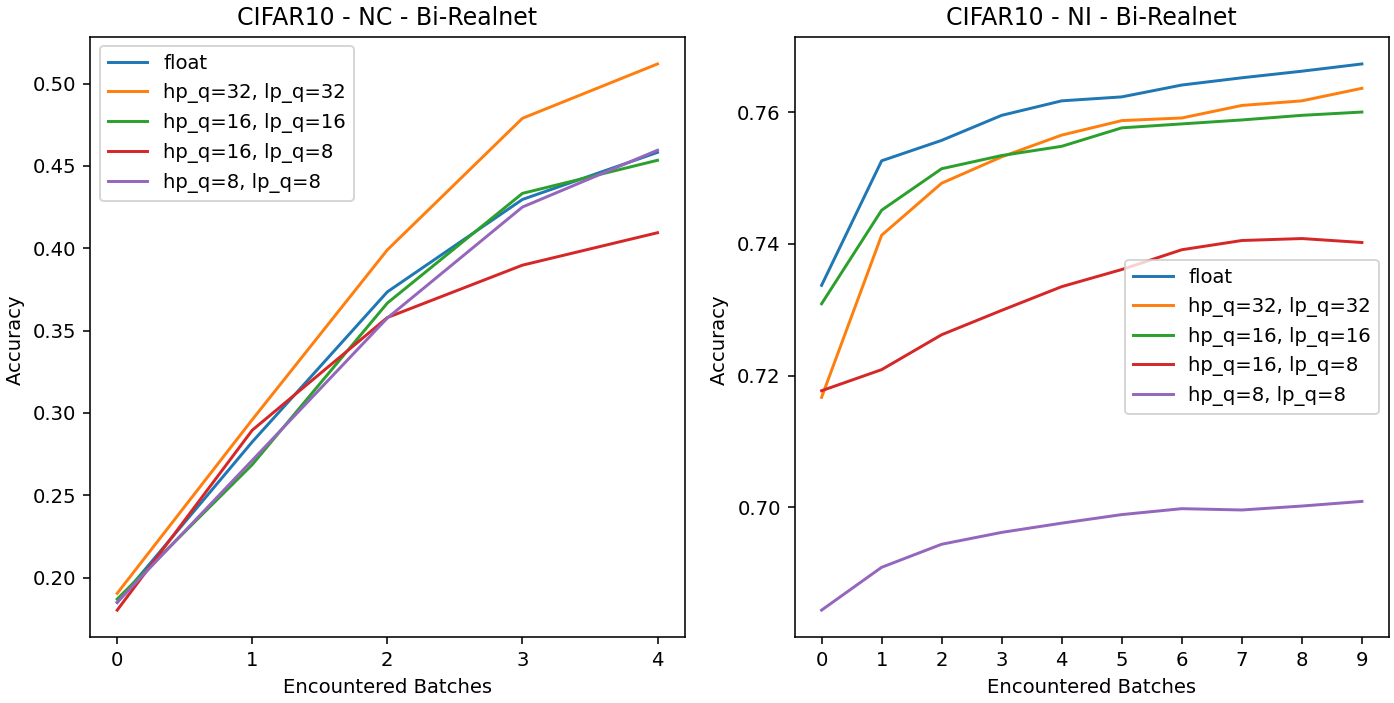}
		\label{fig:bi_realnet_cifar10}
	}
  	\caption{CIFAR10 accuracy results using different quantization methods.}
    \label{CIFAR10_accuracy}
\end{figure}

\begin{figure}
	\centering
	\subfloat[][Reactnet]
	{
		\includegraphics[width=0.55\textwidth]{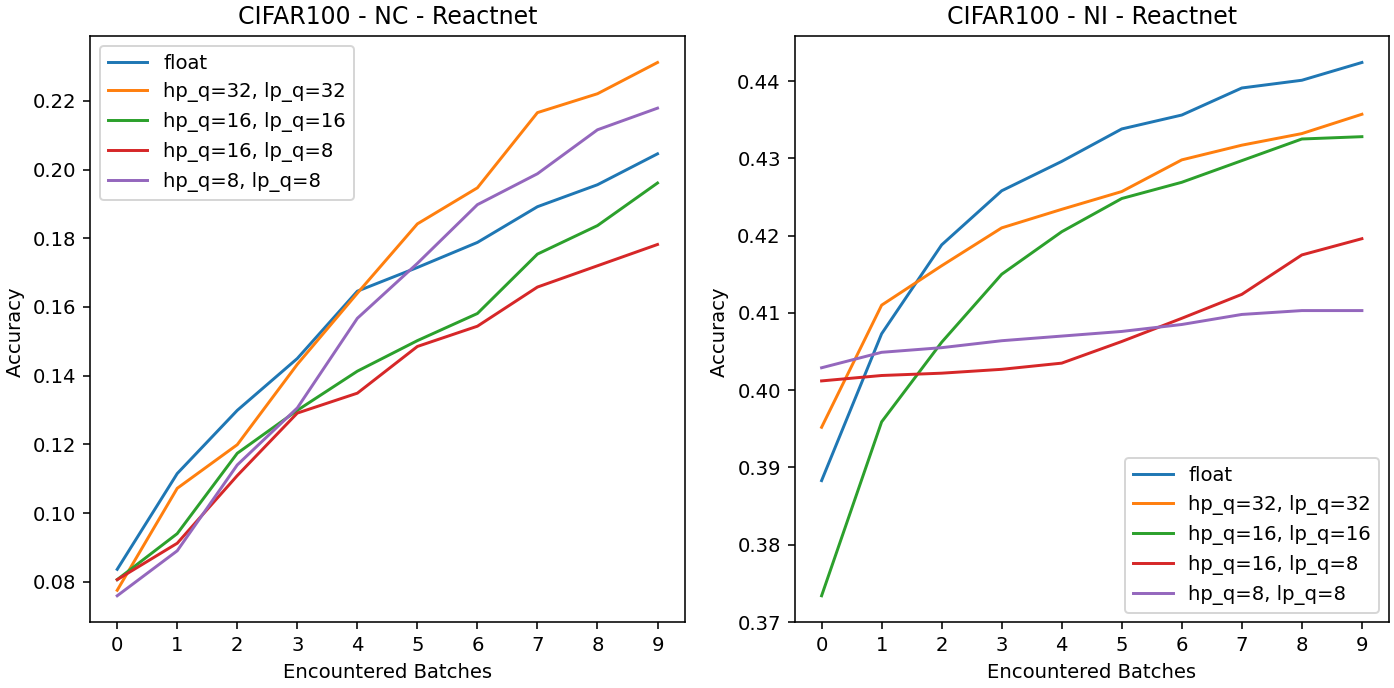}
		\label{fig:reactnet_cifar100}
	}
	\vfill
	\subfloat[][Bi-Realnet]
	{
		\includegraphics[width=0.55\textwidth]{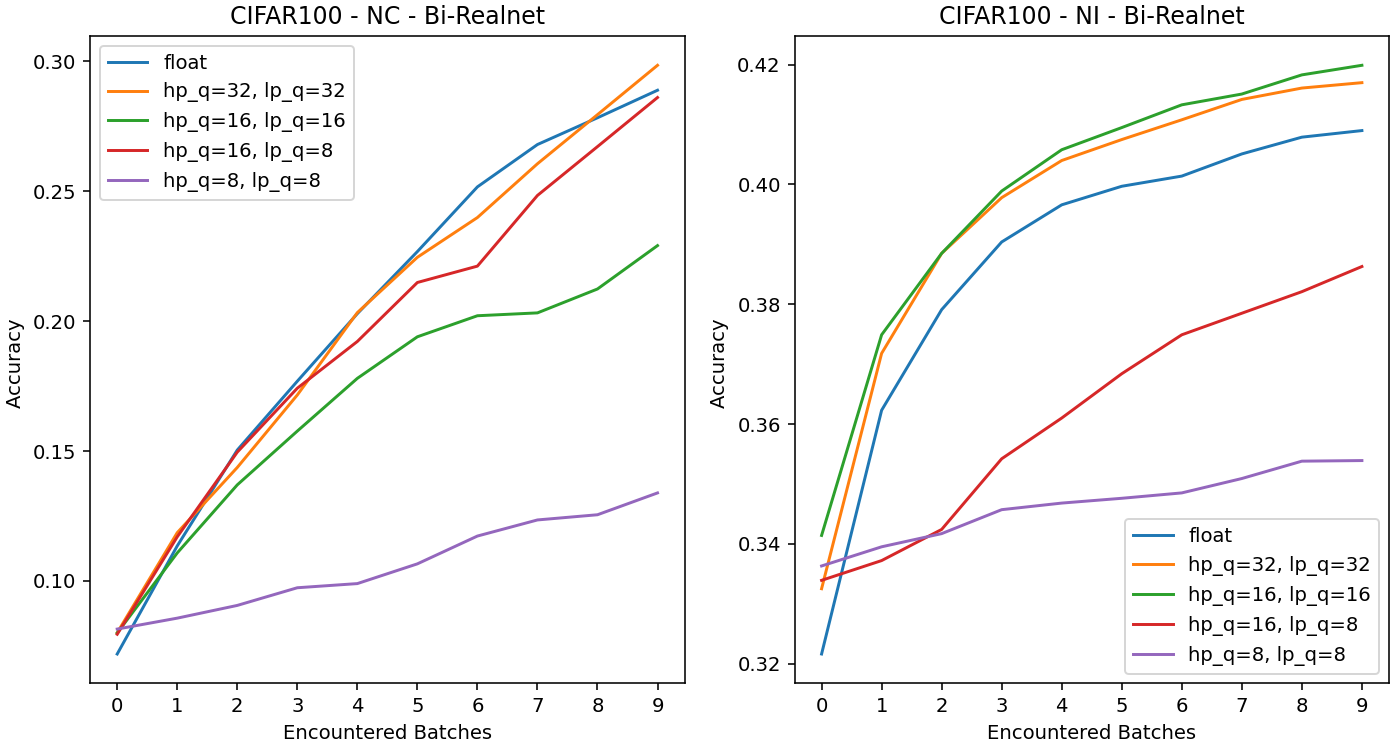}
		\label{fig:bi_realnet_cifar100}
	}
  	\caption{CIFAR100 accuracy results using different quantization methods.}
    \label{CIFAR100_accuracy}
\end{figure}

Our test were performed on both the NI and NC scenario (discussed in Section \ref{continual_learning_section}). Fig. \ref{Core50_accuracy}, \ref{CIFAR10_accuracy} and \ref{CIFAR100_accuracy} summarize the experimental results. On CORe50 dataset (Fig. \ref{Core50_accuracy}) NC scenario, the quantization scheme $lp \textunderscore 8$ gets a consistent accuracy drop over the experiences showing a limited learning capability; instead, the quantizations with $lp \textunderscore 16$ and $lp \textunderscore 32$ reach the same accuracy level of the floating point model. A similar situation can be observed in the NI scenario with the exception of the QuicknetLarge model where the lower quantization schemes are not able to increase the accuracy of the first experience.
For datasets CIFAR10 and CIFAR100 (Fig. \ref{CIFAR10_accuracy} and \ref{CIFAR100_accuracy}) we find similar results for the NI scenario, where the 8-bit quantization scheme limits the learning capability of the model during the experiences. Instead, in the NC scenario, both Bi-Realnet and Reactnet models with $lp \textunderscore 8$ quantization, are able to reach an accuracy result closed to the floating-point model.
From our analysis it appears that the 8-bit quantization of the gradients limits noticeably the learning ability of a binary model when employed in a continual learning scenario for CWR* method. In order to reach accuracy comparable to a floating point implementation we devise the adoption of at least 16 bits both for $lp$ and $hp$; it is worth noting that the computational effort of 16 bits is anyway limited in CWR* because the quantization is confined to the last classification layer.

\section{Conclusion}
\label{resuls_label}

On-device training (or adaptation) can play an essential role in the IoT, enabling the large adoption of deep learning solutions. In this work we focused on implementation of CWR* on edge-devices, relying on binary neural networks as backbone and proposing an ad-hoc quantization scheme. We discovered  that 8-bit quantization degrades too much the learning capability of the model, while 16 bits is a good compromise. To the best of authors' knowledge, this work is the first to explore on-device continual learning with binary network; in the future work we intend to explore the application of binary neural networks in combination of CL methods relying on latent replay \cite{pellegrini2020latent}, which is particularly intriguing given the low memory footprint of 1-bit activation.

\printbibliography

\end{document}